\documentclass[10pt, conference, compsocconf, letterpaper]{IEEEtran}
\IEEEoverridecommandlockouts

\usepackage{cite}
\usepackage{amsmath,amssymb,amsfonts}
\usepackage{algorithm} 
\usepackage{algpseudocode} 
\usepackage{graphicx}
\graphicspath{{images/}}
\usepackage{textcomp}
\usepackage{subfig}
\usepackage{multirow}
 \def\BibTeX{{\rm B\kern-.05em{\sc i\kern-.025em b}\kern-.08em
    T\kern-.1667em\lower.7ex\hbox{E}\kern-.125emX}}
\usepackage{enumitem}
\usepackage{colortbl}
\usepackage{soul}
\usepackage[table,xcdraw]{xcolor}
\usepackage{float}

\newlist{inlineroman}{enumerate*}{1}
\setlist[inlineroman]{itemjoin*={{, and }},afterlabel=~,label=\roman*.}

\newlist{Inlineroman}{enumerate*}{1}
\setlist[Inlineroman]{itemjoin*={{, and }},afterlabel=~,label=\Roman*.}

\begin{document}

\setlist[enumerate]{nosep}
\setlist[itemize]{nosep}
\definecolor{mintgreen}{rgb}{0.6, 1.0, 0.6}
\definecolor{pastelviolet}{rgb}{0.8, 0.6, 0.79}
\definecolor{peridot}{rgb}{0.9, 0.89, 0.0}
\definecolor{richbrilliantlavender}{rgb}{0.95, 0.65, 1.0}
\definecolor{robineggblue}{rgb}{0.0, 0.8, 0.8}

\definecolor{green}{rgb}{0.1,0.1,0.1}
\newcommand{\done}{\cellcolor{teal}done}

\title{\text Enhancing Robotic Navigation: An Evaluation of Single and Multi-Objective Reinforcement Learning Strategies}
\author{\IEEEauthorblockN{Vicki Young$^1$, Jumman Hossain$^2$, Nirmalya Roy$^2$}
\IEEEauthorblockA{$^1$Department of Computer Science, University of San Francisco, San Francisco, USA}
\IEEEauthorblockA{$^2$Department of Information Systems,
University of Maryland, Baltimore County, USA}
\IEEEauthorblockA{$^1$\{vyoung2\}@dons.usfca.edu} $^2$\{jumman.hossain, nroy\}@umbc.edu}

\maketitle
\begin{abstract}

This study presents a comparative analysis between single-objective and multi-objective reinforcement learning methods for training a robot to navigate effectively to an end goal while efficiently avoiding obstacles. Traditional reinforcement learning techniques, namely Deep Q-Network (DQN), Deep Deterministic Policy Gradient (DDPG), and Twin Delayed DDPG (TD3), have been evaluated using the Gazebo simulation framework in a variety of environments with parameters such as random goal and robot starting locations. These methods provide a numerical reward to the robot, offering an indication of action quality in relation to the goal. However, their limitations become apparent in complex settings where multiple, potentially conflicting, objectives are present. To address these limitations, we propose an approach employing Multi-Objective Reinforcement Learning (MORL). By modifying the reward function to return a vector of rewards, each pertaining to a distinct objective, the robot learns a policy that effectively balances the different goals, aiming to achieve a Pareto optimal solution. This comparative study highlights the potential for MORL in complex, dynamic robotic navigation tasks, setting the stage for future investigations into more adaptable and robust robotic behaviors.
\end{abstract}

\section{Introduction}

Robotic navigation, particularly in dynamic and complex environments, is a central problem in the field of robotics and artificial intelligence. With the proliferation of robotics in a myriad of applications - from automated vehicles and industrial automation to healthcare and service sectors, the challenge of effective navigation has garnered significant attention. A critical capability in these applications is the ability of the robot to navigate reliably towards a specific target while avoiding obstacles in its path, thereby ensuring efficient task completion and safety.

\begin{figure}[h]
    \centering
    \includegraphics[width=0.5\textwidth]{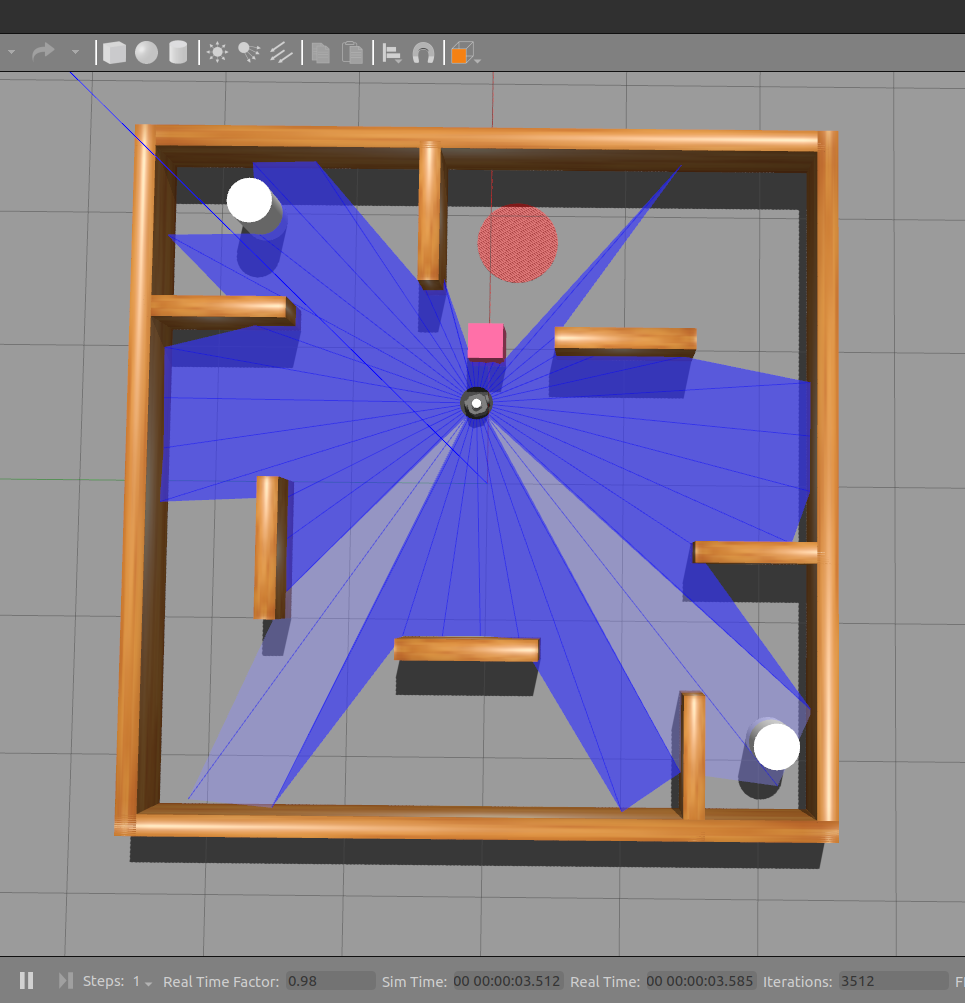}
    \caption{TurtleBot3 robot simulated in a simple environment with the Gazebo framework.}
    \label{figuremesh1}
\end{figure}

Reinforcement Learning (RL) provides a robust framework for training agents to learn from their interactions with the environment and make optimal decisions ~\cite{SuttonRL}. With the advent of deep learning and the development of algorithms capable of handling continuous and high-dimensional action spaces, RL has demonstrated substantial success in various complex tasks, including robotic navigation. Among these, Deep Q-Network (DQN) ~\cite{mnih2015human}, Deep Deterministic Policy Gradient (DDPG) ~\cite{lillicrap2015continuous}, and Twin Delayed DDPG (TD3) ~\cite{fujimoto2018addressing}, have been pivotal, contributing significantly to advancements in the field.

These methods, however, primarily focus on optimizing a single objective function. While this has proven effective in many scenarios, real-world environments often present situations that require balancing multiple, potentially conflicting objectives. For example, in the context of robotic navigation, a robot might need to reach its destination quickly while also minimizing energy consumption and collision risk. The need to address these multifaceted objectives simultaneously has been recognized, leading to the development of Multi-Objective Reinforcement Learning (MORL) \cite{roijers2013survey}.

MORL extends the traditional RL paradigm by optimizing a vector of rewards instead of a single scalar reward, allowing an agent to balance multiple objectives in a more sophisticated manner. While the body of work around MORL has been growing, its applications to complex and dynamic robotic navigation tasks remain relatively unexplored \cite{van2014multi}.

In this paper, we present a comparative analysis of single-objective and multi-objective reinforcement learning strategies for training a robot to navigate efficiently and effectively. Building upon the strengths of MORL, we propose a novel application of this method in the context of robotic navigation. We redefine the reward function to return a vector of rewards, each reflecting a distinct objective, thereby enabling the robot to balance the different goals and aim towards achieving a Pareto optimal solution.

The main contributions of our work are as follows:
\begin{itemize}
    \item A comprehensive comparative analysis between single-objective RL strategies (DQN, DDPG, TD3) and a multi-objective reinforcement learning approach in the context of robotic navigation tasks.
    
    \item A novel application of MORL to complex and dynamic robotic navigation scenarios that inherently involve multiple objectives, providing a method for balancing and simultaneously optimizing these objectives.

    \item An empirical demonstration of the effectiveness of our proposed MORL approach. Through extensive simulations, we show that the proposed approach outperforms traditional single-objective methods, offering more adaptable and robust behaviors across varying levels of environmental complexity.

\end{itemize}

The remainder of this paper is organized as follows: Section 2 reviews related work in the fields of reinforcement learning, single and multi-objective, with an emphasis on their application to robotic navigation. Section 3 defines the problem and presents the methodology employed in our study. The experimental setup and results are detailed in Sections 4 and 5, respectively. The final section, Section 6, concludes the paper and suggests future research directions.

\section{Background Information}

\subsection{Markov Decision Process (MDP)}

In order to use RL to solve a problem, the problem must be structured as a Markov Decision Process (MDP), which consists of five components: an agent, environment, state, action, and reward. The agent is the learner that is to be trained with RL, and it interacts with the environment by taking actions and receiving rewards based on those actions. The environment is the real-world environment with which the agent must operate in and execute actions. States represents the situation of the environment at any point, with different states capturing changes such as the agent's position and direction as well as the position of other objects around it. Actions represent what the agent can do to interact with its environment, where actions may be a defined set of possible action choices or an infinite set. The reward is the positive or negative reinforcement the agent receives from executing a given action in a given state of the environment, and it is the feedback that evaluates the action quality.

In this study, the agent is the robot being trained with RL algorithms within Gazebo environments which have various obstacles. The robot's objective is to navigate to the end goal as efficiently as possible while avoiding obstacles. The robot executes actions by moving, with movements being represented through a continuous, finite range of linear directions and velocities. The state can be captured at any point, and it represents the current position of all objects in the environment (the robot, the obstacles, and the end goal). A penalty is given each time step the robot takes in order to encourage timely navigation, and a large positive or negative reward is given based on the episode outcome: the robot is rewarded for successfully reaching the goal and penalized for failing to, such as by colliding with an obstacle or going beyond the specified time limit. 

An episode is every simulation run which ends with an outcome. Over a sequence of time steps, the agent interacts with its environment through the MDP cycle, doing so until the episode ends, or terminates. By repeatedly training the agent over numerous episodes, the agent will by trial-and-error be able to learn a policy, which is a strategy of deciding what action to take in a given state. RL algorithms therefore aim to train the agent to learn the optimal policy, since the agent will be able to determine what are the best actions to take in any given situation, allowing it to efficiently achieve its objective and maximize the total reward. 



\subsection{Deep Q-Network (DQN)}

While RL algorithms aim to learn the optimal policy, where the optimal action for any given state is known, the optimal policy can be derived indirectly by learning the values of all actions. The value of an action in a given state is called the Q-value. By knowing the Q-values of all possible actions in a given state, the optimal policy can be derived by taking the action with the highest Q-value. Deep Q-Network (DQN) is an RL algorithm which maps a state to the Q-values of all actions that can be taken from that state; it does so through building a table with the Q-values of all actions that can be taken from every state. DQN predicts the Q-values of actions by estimating the Q-function, a process which occurs over training. As DQN is used to train the agent, the feedback that is received is used to update the weights of the Q-function, allowing it to predict the Q-values of actions with greater accuracy. DQN's architecture uses two neural networks, which are optimal function approximators: a Q-network and the target network. The Q-network is the agent that is trained to produce the optimal Q-values. The target network is identical to the Q-network, in that it predicts target Q-values from the next state. In addition, DQN has a component called Experience Replay, which gathers training data by interacting with the environment, allowing this data to be used to train the Q-network. 


\subsection{Deep Deterministic Policy Gradient (DDPG)}

Rather than learning the Q-values of actions, Deep Deterministic Policy Gradient (DDPG) directly learns a policy. While DQN can only predict the Q-values for discrete actions within a finite action setting, where actions are discretely defined, DDPG is advantageous for being able to operate within a continuous action setting ~\cite{lillicrap2015continuous}. For a given state, DDPG directly outputs the best action to take from a continuous action space, hence the reason it is called "deterministic." DDPG is an actor-critic algorithm, meaning it has two neural networks, an actor and a critic. The actor is a policy network, where it takes the state as input and outputs the exact action to take. The critic is a Q-value network, which takes in the state and action and outputs the Q-value. Essentially, the actor chooses what actions to take in a given state, and the critic calculates the predicted Q-value of that action. In addition, the critic and actor each have a target network, which help to stabilize learning. The target networks are used to compute the target Q-values for the next states. Using the feedback received from executing an action in the state, the error between the predicted Q-value and the target Q-value is used to update the actor and critic. 

\subsection{Twin Delayed DDPG (TD3)}

The successor to DDPG, Twin Delayed Deep Deterministic Policy Gradients (TD3) is built to address an issue precedent RL algorithms have with continuously overestimating the Q-values of the critic network ~\cite{fujimoto2018addressing}. A build up of these estimation errors over time causes issues such as the agent falling into a local optima, or extremes (minimum or maximum) of the objective function for a given region of the input space. TD3 focuses instead on reducing overestimation bias. This is done through the addition of three key features: 1) using two critic networks rather than one, 2) delaying updates of the actor network, and 3) adding action noise regularization. For 1), TD3 uses two critic networks to calculate the predicted Q-values and then takes the smaller of the two Q-values, thus favoring underestimation of Q-values. Underestimating values ensures overestimated values are not continuously propagated through the algorithm. 2) deliberately ensures updates to the actor (policy) network are less frequent than updates to the critic (value) network. Training of the agent is badly affected by continuously overestimating poor policies. Delaying updates to the actor allow the critic network to become more stable and reduce errors in estimating Q-values before it is used to update the actor. Lastly, 3) essentially adds noise to selected actions that are fed to the target networks. By doing this, higher target Q-values will be returned for actions that are more robust to noise and interference, essentially giving higher value to more robust actions. This prevents having high variance in the target Q-values that will be used to update the critic networks.

\section{Related Work}


Advancements in the field of reinforcement learning have led to various strategies for optimal control in complex environments. The Q-learning algorithm, as proposed by Watkins and Dayan \cite{watkins1992q}, has become a cornerstone of these strategies. Subsequent improvements led to the development of Deep Q-Network (DQN) by Mnih et al. \cite{mnih2015human}, where deep learning was combined with reinforcement learning for improved performance on complex tasks. For handling high-dimensional, continuous action spaces, Lillicrap et al. \cite{lillicrap2015continuous} introduced the Deep Deterministic Policy Gradient (DDPG) algorithm, which uses deep function approximators. Twin Delayed DDPG (TD3), an improvement over DDPG, was proposed by Fujimoto et al. \cite{fujimoto2018addressing}. It mitigates the issue of overestimation bias by using a pair of Q-functions and delaying policy updates.

However, in scenarios where tasks involve multiple, potentially conflicting objectives, the limitations of the aforementioned single-objective reinforcement learning techniques become evident. The concept of Multi-Objective Reinforcement Learning (MORL) has been explored to tackle this issue, as discussed by Roijers et al. \cite{roijers2013survey}. The applications of MORL in robotic navigation have been the subject of several studies. Van Moffaert and Nowé \cite{van2014multi} presented a scalarized MORL method optimizing energy consumption and travel time, whereas Gomez et al. \cite{gomez2008efficient} employed MORL for optimizing safety and efficiency in robot path planning. Xu et al. \cite{xu2020prediction} proposed an efficient evolutionary learning algorithm to find the Pareto set approximation for continuous robot problems, demonstrating their approach produced denser and higher-quality Pareto policies when balancing between robot movement and energy efficiency. Lastly, Mannion et al. \cite{mannion2021multi} discussed how making decisions under outlined requirements for trustworthy AI is inherently a multi-objective problem. 

More recent works like those of Hwangbo et al. \cite{hwangbo2019learning} and Abdolmaleki et al. \cite{abdolmaleki2018maximum} demonstrate the potential of MORL for efficient robotic navigation and control in real-world scenarios. Hwangbo et al. presented a learning framework that combines model-free and model-based methods, demonstrating its effectiveness in teaching a quadruped robot to run, climb, and jump over obstacles. Meanwhile, Abdolmaleki et al. utilized MORL to optimize performance in diverse simulated physical tasks.

Despite the aforementioned advancements, further exploration is required to fully harness the potential of MORL in complex and dynamic robotic navigation tasks, which is the focus of the current work.

\section{Methods}


\subsection{System Architecture}

In this study, the Gazebo simulation framework is used to simulate testing and training the RL methods to control a TurtleBot3 robot simulation with Robot Operating System (ROS). The TurtleBot3 robot reads data from LiDAR sensors using ROS topics. Additionally, environment information about the end goal location and obstacle locations are also published on ROS topics. ROS nodes such as the DRL Environment Node and the DRL Agent Node subscribe to these topics. ROS robot control features then allow the RL methods to use this sensory and environment information in order to determine how the robot is to navigate toward the end goal. 

\begin{figure}[h]
    \centering
    \includegraphics[width=0.5\textwidth]{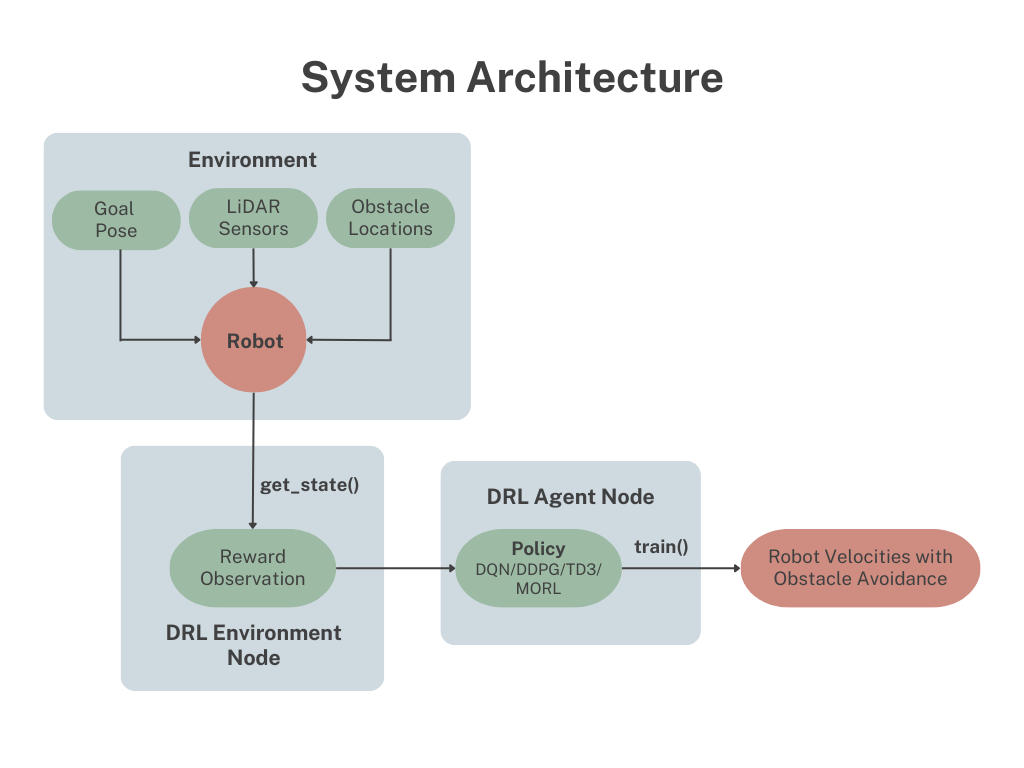}
    \caption{The system architecture describes how information about the environment is input to the reward function, and its output helps the chosen policy decide what action the robot should take.}
    \label{figuremesh1}
\end{figure}

\subsection{Reward Function}

In this study, the single-objective reward function takes in the robot's current action in the form of linear velocity (\verb|action_linear|) and direction of movement (\verb|action_angular|), the robot's distance and angle from the end goal (\verb|goal_dist| and \verb|goal_angle|), its distance from the nearest obstacle (\verb|min_obstacle_dist|), and the status of the robot in terms of reaching the end goal (\verb|succeed|). The reward value is calculated based on the combination of these terms, increasing or decreasing the cumulative result based on how desirable or undesirable the input values are--thereby rewarding or penalizing the behavior of the robot. 








1. Yaw Reward:
\[
r_{\text{yaw}} = -1 \cdot \left| \text{goal\_angle} \right|
\]

2. Angular Velocity:
\[
r_{\text{vangular}} = -1 \cdot \left( \text{action\_angular}^2 \right)
\]

3. Distance Reward:
\[
r_{\text{distance}} = \frac{1}{2} \cdot \left( \frac{2 \cdot \text{goal\_dist\_initial}}{\text{goal\_dist\_initial} + \text{goal\_dist}} - 1 \right)
\]

4. Obstacle Reward:
\[
r_{\text{obstacle}} =
\begin{cases} 
  -20 & \text{if min\_obstacle\_dist} < 0.22 \\
  0 & \text{otherwise}
\end{cases}
\]

5. Linear Velocity Reward:
\[
r_{\text{vlinear}} = -2 \cdot \left( \left( 0.22 - \text{action\_linear} \right) \times 10 \right)^2
\]

6. Total Reward without Success Condition:
\[
\text{reward} = r_{\text{yaw}} + r_{\text{distance}} + r_{\text{obstacle}} + r_{\text{vlinear}} + r_{\text{vangular}} - 1
\]

7. Final Reward with Success Condition:
\[
\text{reward} =
\begin{cases} 
  \text{reward} + 2500, & \text{if succeed = SUCCESS} \\
  \multirow{3}{*}{\text{reward} - 2000,} & \text{if succeed = }\\ 
                                        & \text{COLLISION\_OBSTACLE or }\\
                                        & \text{COLLISION\_WALL} \\
  \text{reward}, & \text{otherwise}
\end{cases}
\]

In step 6, after combining all the individually calculated values into the final reward value, a penalty of \verb|-1| is added to penalize the robot for each time step that the reward function is called, indicating that the robot must reach the end goal quickly in order to avoid the continuous decrease in reward. 

In step 7, a final large reward or penalty may be added to the cumulative reward, depending on the robot's success status: \verb|+2500| is added if the robot successfully reaches the goal (\verb|succeed = SUCCESS|), and \verb|-2000| is added if the robot fails to, namely when it collides with an obstacle (\verb|succeed = COLLISION_OBSTACLE|) or with a wall (\verb|succeed = COLLISION_WALL|). If neither success condition applies, meaning the robot is still navigating to the end goal without success or failure, then no additional value is added.


\subsection{Environments}

The RL methods are evaluated using the Gazebo simulation framework within two different environments: stage A which features three non-moving obstacles and four moving obstacles, and stage B, which features six moving obstacles. The red circle indicates the end goal location which the robot aims to navigate to.

\begin{figure}[h]
    \centering
    \includegraphics[width=0.5\textwidth]{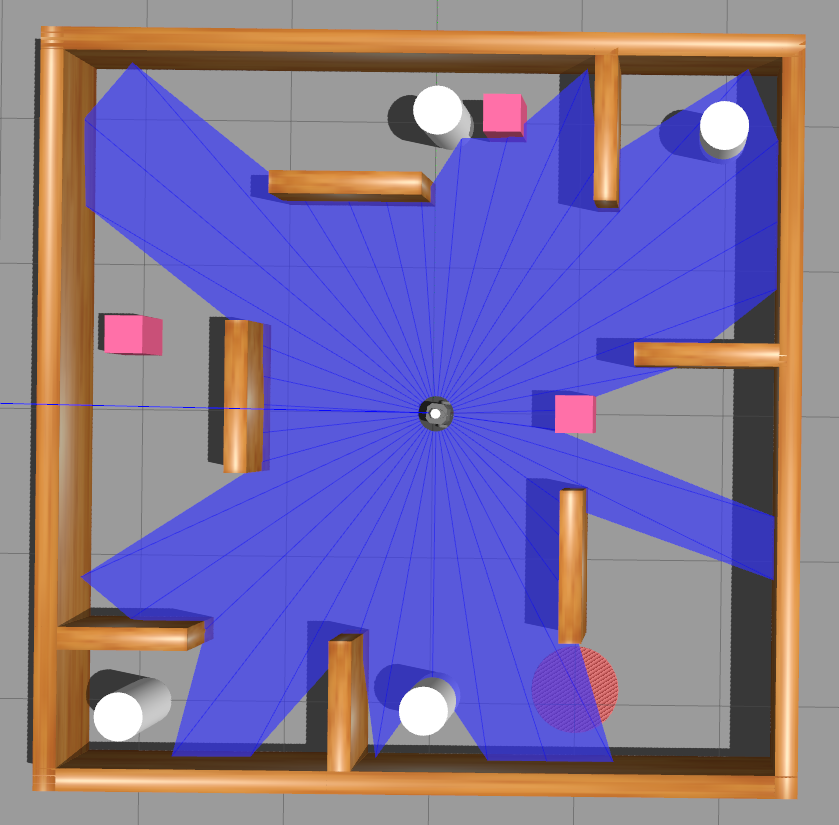}
    \caption{Stage A has an environment with three non-moving obstacles (pink) and four moving obstacles (white).}
    \label{figuremesh1}
\end{figure}

\begin{figure}[h]
    \centering
    \includegraphics[width=0.5\textwidth]{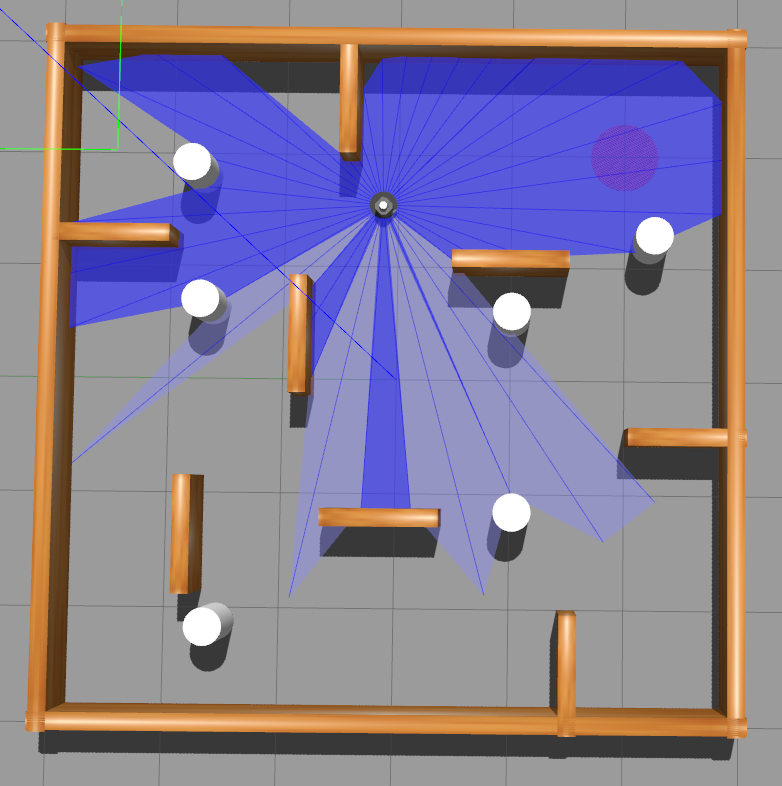}
    \caption{Stage B has an environment with six moving obstacles (white).}
    \label{figuremesh1}
\end{figure}

\section{Results}

For training in both stages A and B, DDPG performed the best in terms of having episodes yielding the highest rewards, followed by TD3, then DQN. All three methods produced overall higher reward scores when training in stage B than in stage A, as stage A is more difficult to navigate than stage B due to having non-moving obstacles blocking key pathways of the environment. However, all three algorithms still demonstrated consistent collisions with moving and non-moving obstacles in both stages A and B. 

For stage A, DDPG's best performing episodes were demonstrated in later episodes of training, with top four episodes \verb|[3200, 3300, 3500, 3000]| returning reward scores \verb|[-2322, -2349, -2484, -2551]| respectively. 

In contrast, TD3's performance in stage A demonstrated better performing episodes in earlier episodes of training, with top four episodes \verb|[1200, 900, 600, 1000]| returning reward scores \verb|[-2517, -2892, -2938, -2943]| respectively. All of TD3's best performing episodes had reward scores that were lower than that of DDPG's. 

Lastly, the stage A performance of DQN showed better performing episodes in the middle of training, with top four episodes \verb|[2700, 2800, 3200, 2000]| returning reward scores of \verb|[-3579, -3766, -3782, -3823]| respectively. All of DQN's best performing episodes had the lowest reward scores of the three algorithms in stage A.

\begin{figure}[h]
    \centering
    \includegraphics[width=0.5\textwidth]{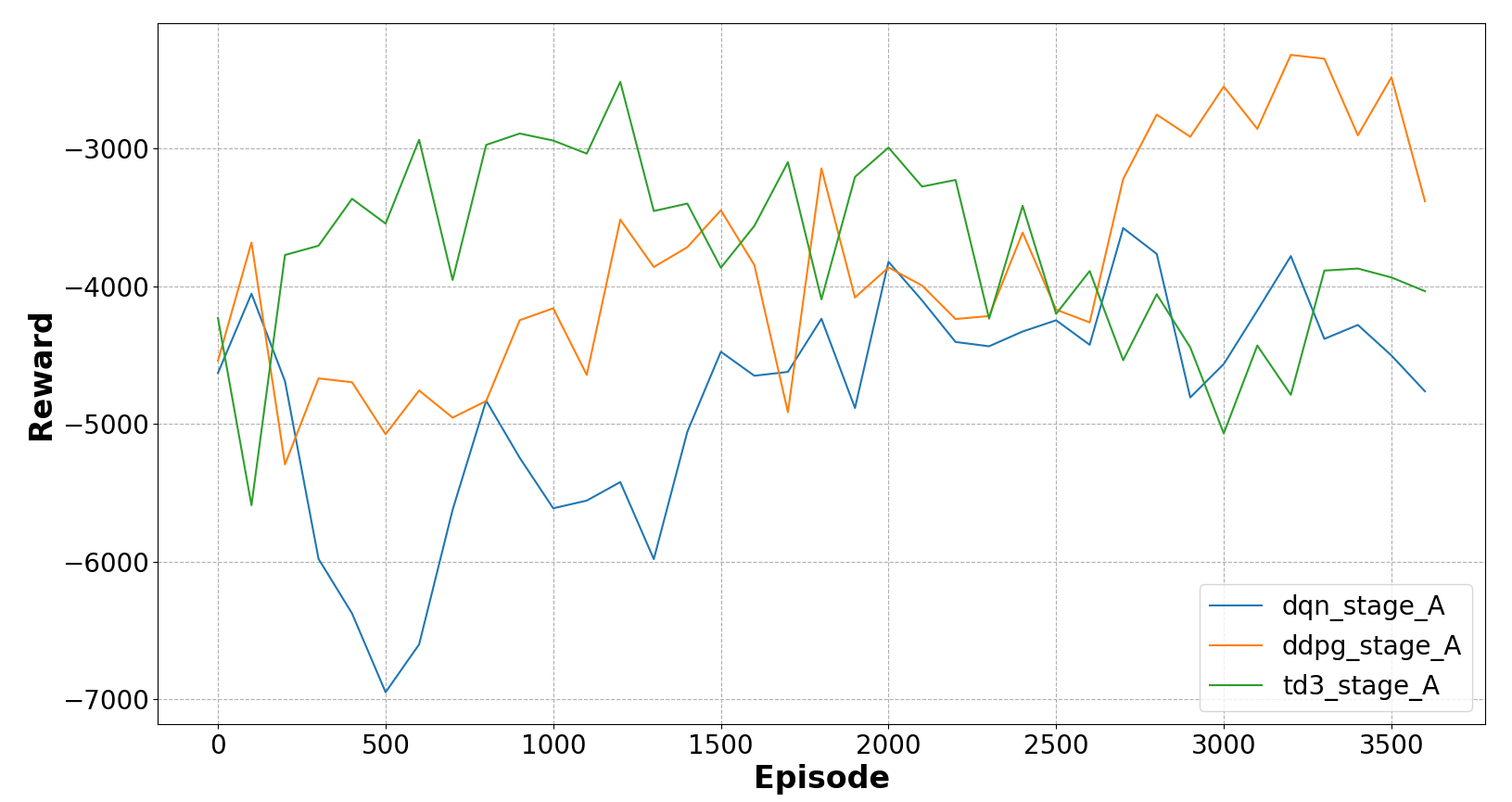}
    \caption{The reward graph shows the performance of DQN, DDPG, and TD3 in stage A over the course of 4000 episodes.}
    \label{figuremesh1}
\end{figure}

For stage B, DDPG again outperformed DQN and TD3, with best performing episodes witnessed near the later part of training: the top four best performing episodes \verb|[2600, 2800, 2900, 3200]| returned reward scores \verb|[-1126, -1509, -1512, -1522]| respectively. 

TD3's performance in stage B also demonstrated the best performing episodes in the later episodes of training, with top four episodes \verb|[3600, 3500, 3300, 3400]| returning reward scores \verb|[-1637, -1766, 2064, -2103]| respectively. The top reward scores from TD3 in stage B were lower than the top reward scores from DDPG.

Lastly, DQN's best performing episodes from stage B were spread out from the middle to the end of the training duration, with top four episodes \verb|[2000, 3500, 3100, 3600]| returning reward scores \verb|[-2401, -2532, -2578, -2629]| respectively. DQN's top performing episodes in stage B ultimately returned the lowest reward scores of the three algorithms. 

\begin{figure}[h]
    \centering
    \includegraphics[width=0.5\textwidth]{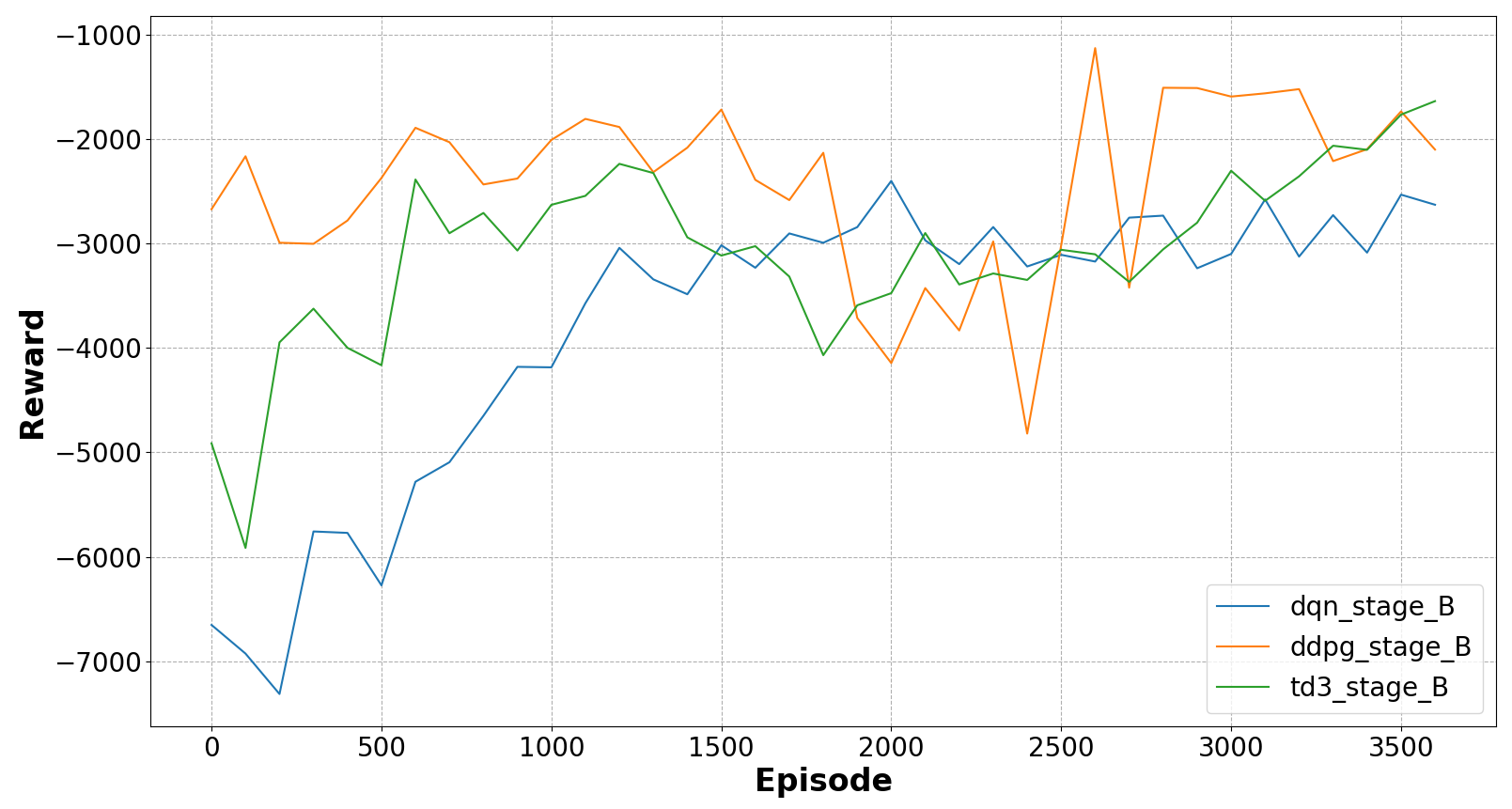}
    \caption{The reward graph shows the performance of DQN, DDPG, and TD3 in stage B over the course of 4000 episodes.}
    \label{figuremesh1}
\end{figure}

Fig. 5 and Fig. 6 show the reward graphs for the performance of DQN, DDPG, and TD3, in stage A and B respectively. These reward graphs demonstrate that DDPG generally returned the highest rewards over the course of 4000 episodes. DQN is noted to consistently return the lowest reward scores in both stages. However, while DDPG generally outperformed TD3 in training, TD3 had a more stable learning curve that may demonstrate greater improvement if testing beyond 4000 episodes. This is likely due to TD3’s structure which reduces overestimation of action values. Because DDPG's structure does not prevent against this overestimation bias, this may account for the volatility witnessed in DDPG's returned reward scores. For future implementation of MORL algorithms, TD3 could be modified to accept a vector of reward values from an MORL reward function. 

While the individual performance of all algorithms were recorded, only DDPG's performance in stage A is shown as an example in Fig 7. For DDPG's individual performance in stage A, it is observed that the average actor loss gradually decreased, meaning the actor generally learned the optimal robot actions, though not at a consistent rate, as indicated by the spikes in the graph. On the other hand, critic loss remained relatively high, indicating that the critic failed to estimate the Q-values of these actions. Additionally, there was a total of more collisions with walls and moving obstacles than there was of successful episodes reaching the goal. The average reward also did not increase significantly over episodes. Although only DDPG's performance for stage A is shown, the individual performances of the other algorithms in both stages A and B demonstrated similar issues with  actor and critic loss not minimizing over the course of training, relatively low average rewards, and consistently high collisions with walls and moving obstacles.

\begin{figure}[h]
    \centering
    \includegraphics[width=0.5\textwidth]{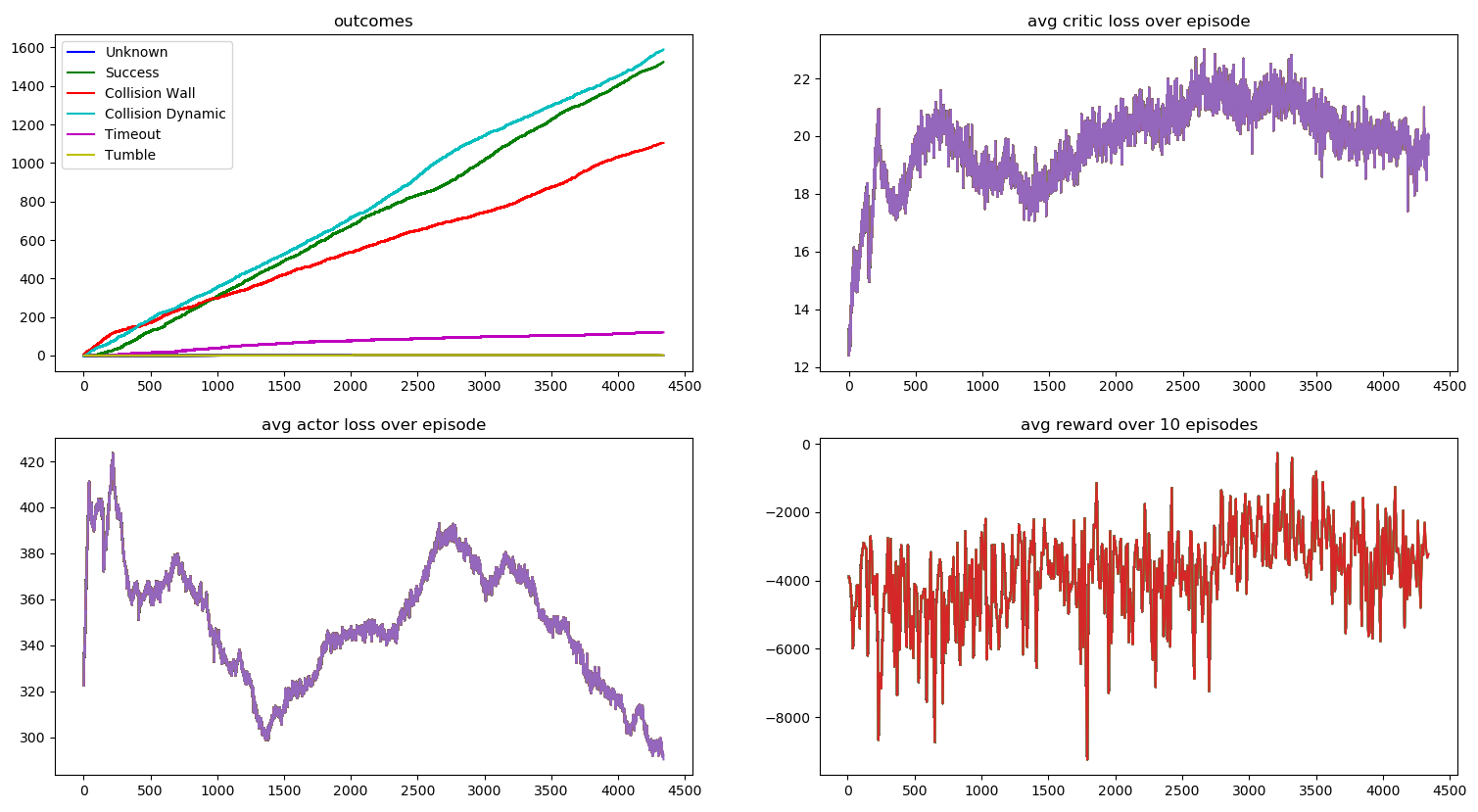}
    \caption{The performance of DDPG in stage A over 4000 episodes is shown.}
    \label{figuremesh1}
\end{figure}

Overall, DQN, DDPG, and TD3 did not perform optimally with consistent success to reach the goal, particularly so when the end goal was located in more difficult locations, such as the corners of the environment, where there were more obstacles in the way. Frequent collisions with obstacles continued to be observed even after constant training, and the robot did not reach the goal at optimal speeds. While the algorithms gradually learned to reach the end goal and therefore optimize the robot's single objective with relative success, they did so at the cost of neglecting other priorities, such as avoiding obstacles and finding the shortest and quickest path to the end goal. These algorithms consequently demonstrated overall poor performance, however these results are to be expected, and in fact demonstrate the main issue: single-objective RL strategies are unsuitable for providing a solution to the problem of autonomous robotic navigation. 

Improvements could be made with the implementation of an MORL algorithm and reward function in order to optimize multiple objectives. DQN, DDPG, and TD3 demonstrated a failure to minimize collisions and reach the goal in a quicker manner, and the single-objective reward function returned a reward score that was composed of a large reward for successfully reaching the goal or a large penalty for failing to. This large reward/penalty encouraged solely prioritizing the objective at the cost of neglecting other priorities, as the smaller rewards/penalties the robot gained for its movements, distance to the goal, and distance to other obstacles were a minuscule component of the cumulative reward. If the robot was able to reach the goal location despite taking the longest path or navigating narrowly close to obstacles, it was encouraged to do so by the large reward that it gained from successfully doing so. 

Consequently, the suitable solution is an MORL algorithm which uses a modified reward function that returns a vector of rewards rather than a single reward. This reward function allows a way to simultaneously balance and optimize having multiple objectives, as each objective is represented by its own reward score, therefore enabling the rewarding or penalizing of particular behaviors in regards to each objective. This also allows modifying the reward function and the modeled behavior of the robot according to preferences for particular objectives. For example, a robot with safer navigation behavior can be preferred by choosing behaviors which earn higher reward values for collision avoidance.

\section{Conclusion and Future Directions}


We presented a comprehensive analysis of the performances between single-objective RL strategies (DQN, DDPG, TD3) and highlighted the issues with using such strategies for the problem of robot navigation, and examined how such issues could be resolved through using multi-objective RL strategies instead. We ran extensive training simulations with each single-objective algorithm and examined the individual performances of each one in two different environments, before comparing their performances through generated reward graphs. We demonstrated how the performance of single-objective RL strategies suffer as a result of being unable to provide an optimal solution which balances and simultaneously optimizes multiple objectives. 

In the future, we would like to test the trained models in order to fully demonstrate the performance of each algorithm, in addition to finding ways to increase the success rate of performance. Moreover, we plan to implement and test multi-objective RL (MORL) strategies. Recent advancements in MORL have shown promising results in resolving issues related to conflicting objectives, a limitation inherent in single-objective RL strategies. By applying MORL, we aim to provide a more holistic solution to the complex problem of robotic navigation ~\cite{hossain2023covernav}.

Additionally, we intend to extend our research to physical robots for further testing and training. This step is crucial for understanding how these algorithms perform in real-world scenarios, as opposed to controlled, simulated environments ~\cite{spector2022simulated}. By incorporating SynchroSim's insights ~\cite{dey2022synchrosim}, we plan to focus on synchronization and communication aspects of single-objective and multi-objective RL strategies in multi-robot systems. The comparison of single-objective and multi-objective RL strategies on different physical robots will also provide valuable insights into the adaptability and practicality of these approaches in varied robotic platforms.

We also plan to load the models onto a physical robot for further testing and training, in addition to comparing the performance of single-objective and multi-objective RL strategies on different physical robots as well. 





\bibliographystyle{unsrt}
\bibliography{bibliography}


\end{document}